\documentclass{article}
\usepackage{spconf,amsmath,graphicx}
\usepackage{amssymb}
\usepackage{array}
\usepackage{booktabs}
\usepackage{multirow}
\usepackage{algorithm}
\usepackage[noend]{algpseudocode}
\usepackage{hyperref}

\def\O{{\cal O}}
\def\W{{\cal W}}
\DeclareMathOperator*{\argmax}{arg\,max}

\makeatletter
\def\BState{\State\hskip-\ALG@thistlm}
\makeatother

\title{DISCRIMINATIVE TRAINING OF RNNLMS WITH THE AVERAGE WORD ERROR CRITERION}
\name{R\'emi Francis, Tom Ash, Will Williams}
\address{Speechmatics, Cambridge, UK\\\texttt{\{remi,toma,willw\}@speechmatics.com}}

\begin{document}
%
\maketitle
\begin{abstract}
In automatic speech recognition (ASR), recurrent neural language models (RNNLM) are typically used to refine hypotheses in the form of lattices or n-best lists, which are generated by a beam search decoder with a weaker language model. The RNNLMs are usually trained generatively using the perplexity (PPL) criterion on large corpora of grammatically correct text. However, the hypotheses are noisy, and the RNNLM doesn't always make the choices that minimise the metric we optimise for, the word error rate (WER). To address this mismatch we propose to use a task specific loss to train an RNNLM to discriminate between multiple hypotheses within lattice rescoring scenario. By fine-tuning the RNNLM on lattices with the average edit distance loss, we show that we obtain a 1.9\% relative improvement in word error rate over a purely generatively trained model.
\end{abstract}
\begin{keywords}
Discriminative training, recurrent neural network language model, average word error rate, lattice rescoring, automatic speech recognition
\end{keywords}
\section{Introduction}
\label{sec:intro}

\textit{Automatic speech recognition} (ASR) systems are made from two majors components, the \textit{acoustic model} (AM), and the \textit{language model} (LM). The AM computes the probability of the acoustic observations given a sentence, and the LM encodes a prior over possible sentences. 

\begin{equation} \label{eq:decoding}
\hat{\W} = \argmax_\W \mathbb { P }(\W | \O) = \argmax_\W \mathbb { P }(\W) \cdot \mathbb { P }( \O | \W)
\end{equation} 
Where $\O$ is the acoustic observation, $\W$ spans all the possible sentences. \cite{Yu:2014:ASR:2695502}

The Equation (\ref{eq:decoding}) is not tractable as the set of sentences is infinite. In practice, we restrict the search space with a beam search using a \textit{weighted finite state transducer} (WFST) created from a small n-gram \cite{Mohri2008}. The output of the search is a list of hypotheses, in the form of an n-best list or a lattice. The hypotheses of this list are then rescored with a stronger language model, such as a bigger n-gram, or a neural network.  
The LMs are trained separately from the acoustic model, in an unsupervised way to minimise the negative log-likelihood of the data, or equivalently, the \textit{perplexity} (PPL).
Given $\W = w_1 \dots w_T$ a sentence, its probability is computed with the joint probability:
\begin{equation}\label{eq:lmprob}
\mathbb { P } \left( \W \right) = \prod _ { t = 1 } ^ { T } \mathbb { P } \left( w_t | w_{t-1} , \dots , w_{1} \right)
\end{equation} 
In this formulation, the probability of a word depends on the previous words, and in the case of an n-gram, the history use is truncated to the $n-1$ preceding words. This makes n-grams fast and suited for short sentences and decoding, but they can't model long range dependencies. With RNNLMs the history is theoretically infinite because it's encoded in a continuous space, however they require a lot more compute, so they can only be used to discriminate between the hypotheses generated by the decoder. 

Most of the literature is focusing on minimising the PPL of the models, without considering their discriminative power on the tasks they are applied to, like ASR or machine translation. In ASR, the quality of a transcription is evaluated with the \textit{word error rate} (WER), based on the Levenshtein distance of the transcript against the reference.

Although PPL has initially been shown to correlate with WER \cite{Klakow2002}, training the RNNLM to minimise the PPL is effectively using a surrogate loss for the rescoring task, and different training schemes can yield models that perform well for WER but poorly for PPL \cite{Huang}.  Moreover, the hypotheses generated by the decoder contain noise from the decoding process, of a form the RNNLM would not have seen in its training data. Additionally, because the language models scores being a sum of log-probabilities, longer sentences tend to have lower scores than shorter ones, which would encourage deletions.

The goal of this paper is to address these shortcomings and tune the RNNLM with a discriminative loss so it can learn to discriminate between noisy hypotheses and give a better score to the ones that are likely to minimise the WER. Unlike \cite{Huang}, we chose to evaluate the discriminative techniques that have been proven to work for acoustic model training.

\section{Related work}
\label{sec:related work}

\subsection{Acoustic models}
\label{ssec:dam}
Discriminative training for AMs has been developed first for \textit{gaussian mixture models} (GMMs), but has then been successfully applied to neural networks. \cite{Yu:2014:ASR:2695502}. These sequence discriminative criteria compute probabilities over all possible decoding paths, which is not tractable. Thus they are approximated with lattices and alignments obtained from decoding the training data with a beam search, so the models have to be pre-trained with another objective function first, usually \textit{cross-entropy} (CE).

\cite{Povey2013} evaluates many different criteria from the families of \textit{maximum mutual information} (MMI) and \textit{level minimum bayes risk} (MBR); and \textit{state-level minimum bayes risk} (sMBR) is found to produce the best model.
In \cite{Povey2016}, a \textit{lattice-free} version of MMI (LF-MMI) is used the train the neural networks, relaxing the need to pre-train with cross entropy, and they show that the network can still be improved using sMBR. In \cite{Shannon2017} a word level MBR objective is defined, called \textit{edit-based MBR} (EMBR), which brings the objective function even closer to the WER, and provides improvements over sMBR.


\subsection{Language models}
\label{ssec:dlm}
There is some literature about discriminative training of language models, but most of it pre-dates the use of neural networks in language modelling, and use engineered features to train a perceptron or a SVM classifier \cite{Dikici2013}. 

More recently, \cite{Tachioka2015} presented an RNNLM to discriminate between the reference transcript and the one-best hypothesis by optimising the differences in their cross-entropies. The WER improvement they report is marginal, and \cite{Huang} provide an improvement of their loss function by introducing a margin loss, which they call \textit{Large Margin Language Model}:
\begin{equation}\label{eq:margin}
\sum_{j = 1}^{n} \max \left\{0, \tau - \left( \log \mathbb{P} \left( \W_{\mathrm{ref}} \right) - \log \mathbb { P } \left( \mathbf \W_{\mathrm{hyp}_{j}} \right) \right) \right\}
\end{equation}
Where $n=256$ is the number of candidates in an n-best list. This encourages the score of the reference to be greater than the hypotheses' by $\tau$. They also propose a similar objective where the candidates are ranked with each other such that the ones with lower WER are given a better score. These losses are used to fine-tune an RNNLM, and they show a more substantial gain in WER. They also note that the PPL of the models produced by the fine-tuning is greatly degraded.

This large margin loss is similar to the contrastive entropy loss introduced by \cite{Arora2016a}, which aims to quantify the ability of a language model to discriminate between a sentence and artificially noised versions of it. They show that a model trained with this loss has a better discriminative power, however no ASR results are presented.


\section{Minimum word error loss}
\label{sec:loss}
\subsection{Formulation}
\label{ssec:formulation}
The aim of this work is to train the RNNLM to assign scores to the lattice arcs to minimise the true metric of interest, which for ASR is the WER. 

We wish to apply the discriminative training techniques that have been proven to work with acoustic models. We chose to evaluate MBR training, which minimises the expected loss over all the paths $\pi$ in the lattice:
\begin{equation}\label{eq:embr}
\mathbb{E}_\pi \left[ L \left(\W, \W_{\mathrm{ref}} \right) \right] = \sum_{\pi} \mathbb{P}(\pi | x, \lambda ) L \left( \W_{\mathrm{hyp}}(\pi), \W_{\mathrm{ref}} \right)
\end{equation}
Where $L(\W, \W_{\mathrm{ref}})$ can be any loss measuring the distance between the reference $\W_{\mathrm{ref}}$ and a hypothesis $\W$, and $\mathbb{P}(\pi | x, \lambda )$ is the probability of $\pi$ given the acoustic input and the model parameters.

We use the same loss as in \cite{Shannon2017}, the EMBR. In this formulation $L$ is the word level edit distance, the same as the one used to compute the WER. 
Moreover, since the RNNLM operates on words in the lattice, rather than acoustic states, EMBR appears to be more appropriate than MPE or sMBR.

\subsection{Computation of the loss}
\label{ssec:computation}
The WER doesn't decompose additively over the arcs of the lattice \cite{Heigold2005,VanDalen2015}, so it doesn't obey the ring composition rules usually used in the FST algorithms. This makes computing Equation (\ref{eq:embr}) impractical when lattices are generated on the fly during acoustic model training. \cite{Shannon2017} approximates this expectation by sampling $100$ paths from the lattice, and in \cite{Prabhavalkar2017} the n-best paths are selected instead. In \cite{Povey2002}, another approximation of the WER contribution of each arc is formulated by using a time alignment of the lattice with the reference.

In these papers, the use of an approximation is motivated by the need of a fast way to compute the loss. However, in our training regime, the acoustic model is fixed, and the lattices generated by the decoder don't depend on the RNNLM, therefore we pre-compute and process them to obtain the exact WER information for each arc using the algorithm described in \cite{Heigold2005}. This allows us to avoid using an approximation, and in practice we observe that the lattices with WER information are on average only 20$\%$ larger than the original ones. The EMBR can then be computed with the same forward-backward algorithm as described in \cite{Povey2002}.

\subsection{Lattice rescoring}
\label{ssec:rescoring}
Rescoring the lattice involves changing the scores present on its arcs. They are generated with an FST created from a small n-gram, usually a pruned 3-gram or 4-gram. When rescoring a lattice with a more powerful model, some states need to be expanded to account for the bigger history used by the language model. RNNLMs have a theoretically infinite history, and would therefore require expansion of the lattice into a tree, which is not practically tractable, because the number of possible paths increases exponentially with the length of the utterance, so approximations have to be used.  

Many different ways of pruning the paths are possible \cite{Liu2016,Kumar2017,Xu2018}. We choose to expand the lattice with a 4-gram language model, and where a state has multiple incoming arcs, we chose the history corresponding to the path with the best score.

\subsection{Initialisation and training}
\label{ssec:initialisation}
Before training with the EMBR loss, we initialise the RNNLM by training it conventionally on large corpora. Instead of cross entropy we use the \textit{noise contrastive estimation} (NCE) loss, because it's much faster to train, and yields a self-normalised model which doesn't require evaluation of a softmax over the whole vocabulary for each arc in the lattice \cite{Mnih2013,Williams2015}.  
The reason for this pre-training is because the RNNLM is only effective when trained on large amounts of data; this typically amounts to orders of magnitude more data than is available from reference transcripts.

It has been often found in the literature of discriminative AM training \cite{Prabhavalkar2017,Povey2002} that interpolating with the cross entropy loss helps stabilise the training and prevent over-fitting, so we experiment with adding NCE to the EMBR loss, where the NCE loss is computed on the reference transcript.
\section{Experimental setup}
\label{sec:setup}
\subsection{Data}
\label{ssec:data}
The language model data used to pre-train the model is 2.5 billion words of general English text and the acoustic modelling data is 2000 hours of transcribed general English, about 900k utterances (21 million words), which is perturbed with point-source and reverberant noise \cite{Ko2017}.   

We compute our test results on different test sets representing different domains: news, podcast, radio, entertainment, meetings, and political; each of them having a duration of 4 hours, which is between 35,000 and 45,000 words. They have been chosen to be representative of the variability in accents, audio difficulty and language that could be encountered in a real scenario.

\subsection{Model description}
\label{ssec:description}
The training and decoding of acoustic models was performed with Kaldi, using the LF-MMI objective and the \textit{time delay neural network} architecture similar to the one described in \cite{Povey2016}, and we used PyKaldi \cite{pykaldi} to process the lattices in Python. 

We trained a 4-gram on the language model data, and pruned it to 60 million n-grams. We use it to expand the lattice, and in all results involving the RNNLM, we interpolate the RNNLM probabilities with the 4-gram, with a weight of 0.9 for the RNNLM and 0.1 for 4-gram.

The RNNLM was trained with NCE using TensorFlow and is a \textit{gated recurrent unit} (GRU), with a single hidden layer of size 512, and vocabulary size of 125,000 words. We trained this model with \textit{stochastic gradient descent} (SGD) for 15 epochs, decaying the learning by 4 when the PPL on the validation set didn't improve by more than 1 after each epoch. We then selected the model that has the best PPL on the validation set. 

The EMBR fine-tuning was also performed in TensorFlow, with a fixed learning rate of 0.01 with SGD. We batched lattices by number of states to make the processing more efficient, and we used a batch size of 32. To interpolate with the NCE loss, we used the reference transcript of the lattices used for the EMBR loss.

In order to verify that the source of the improvements are not due to adapting the RNNLM to the acoustic data's language domain, we also perform two additional fine-tuning experiments. In the first we adapt the RNNLM on the transcripts of the acoustic data, and in the second, we adapt it on the oracle transcripts we get from decoding the acoustic data.

\begin{table*}[!htpb]
\centering
\begin{tabular}{c|ccccccc|c}
\toprule
\rule[-1ex]{0pt}{2.5ex} \multirow{2}{*}{Model} & \multicolumn{8}{c}{WER(\%) / Relative improvements compared to 4-gram only (\%)}\\ 
\cline{2-9}
\rule[-1ex]{0pt}{2.5ex}  & Radio & Podcast & News 1 & News 2 & Media 1 & Media 2 & Meetings & Average \\ 
\midrule
\rule[-1ex]{0pt}{2.5ex} oracle & 4.8 & 5.2 & 8.8 & 9.8 & 11.8 & 20.9 & 33.9 & 13.6  \\ 
\midrule
\rule[-1ex]{0pt}{2.5ex} 4-gram only & 11.3 & 10.5 & 13.7 & 17.2 & 25.5 & 36.8 & 47.3 & 23.2  \\ 
\rule[-1ex]{0pt}{2.5ex} RNNLM & 9.9 / 12.4 & 9.0 / 14 & 12.6 / 7.8 & 15.8 / 7.9 & 22.8 / 10.6 & 34.8 / 5.4 & 45.4 / 4.0 & 21.5 / 7.3 \\ 
\rule[-1ex]{0pt}{2.5ex} RNNLM + EMBR & \textbf{9.4 / 16.9} & \textbf{8.6 / 17.4} & \textbf{12.5 / 8.6} & \textbf{15.7 / 8.8} & \textbf{22.4 / 12.2} & \textbf{34.5 / 6.2} & \textbf{44.9 / 5.2} & \textbf{21.1 / 9.1} \\
\midrule
\rule[-1ex]{0pt}{2.5ex} Adapting on transcripts & 9.8 & 8.8 & \textbf{12.5} & 15.9 & 22.7 & 35.1 & 45.2 & 21.4  \\
\rule[-1ex]{0pt}{2.5ex} Adapting on oracles & 10.0 & 8.9 & \textbf{12.5} & 15.9 & 22.7 & 35.2 & 45.2 & 21.5  \\  
\bottomrule
\end{tabular} 
\caption{WERs on the test sets.}
\label{table:wer}
\end{table*}
\section{Results}
\label{sec:Results}
\begin{figure}
  \centering
  \def\svgwidth{\columnwidth}
  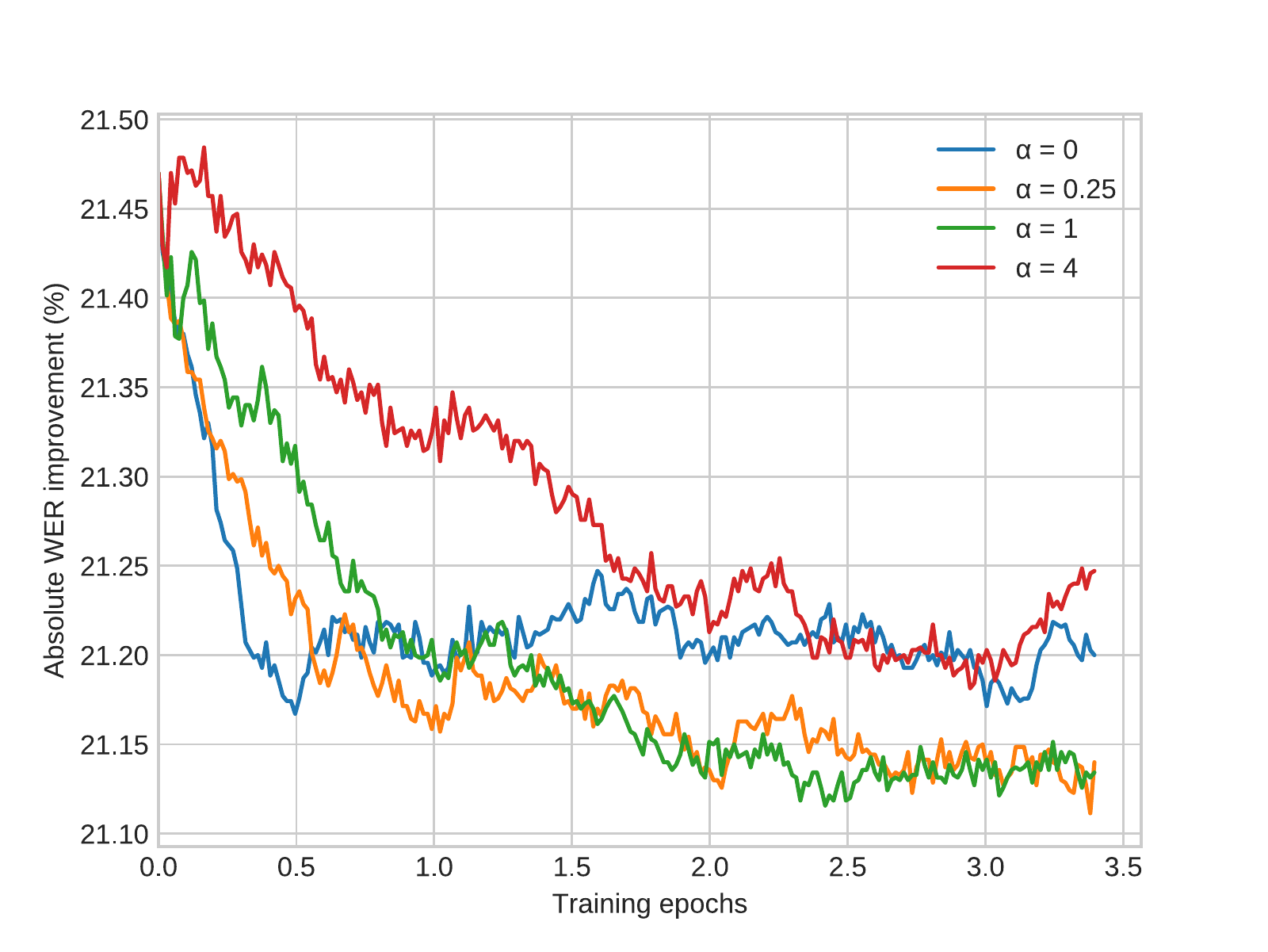
  \caption{WER on the test set, measuring the impact of the factor $\alpha$ in the total loss: $\mathcal { L } _ { total } = \mathcal { L } _ { \mathrm {EMBR} } + \alpha \mathcal { L } _ { \mathrm { NCE } }$ }
  \label{fig:alpha}
\end{figure}
\begin{figure}
  \centering
  \def\svgwidth{\columnwidth}
  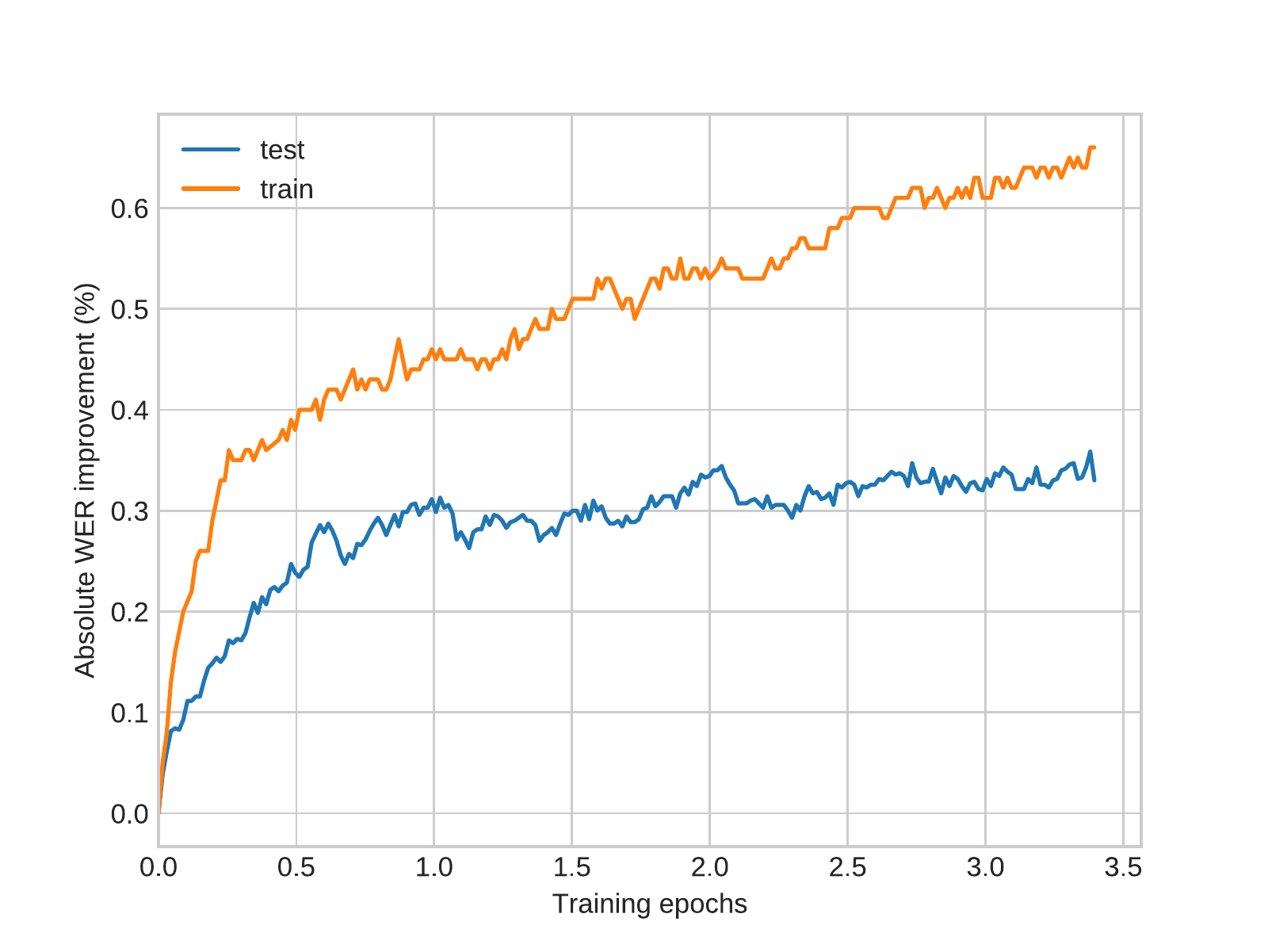
  \caption{Evolution of the absolute improvement in train and test WER during training for $\alpha=0.25$.}
  \label{fig:train_test}
\end{figure}
We first evaluate the impact of interpolating with the NCE loss. Figure \ref{fig:alpha} illustrates the sensitivity of the training to this parameter, and demonstrate that a low but non-negative value of $\alpha$ is required to prevent over-fitting. The best model was produced with $\alpha = 0.25$.

In Table \ref{table:wer} we compare the WER of the baseline RNNLM with the RNNLM+EMBR, and the adaptation experiments. We also report the oracle WER because it's a lower bound for the WER after rescoring since this operation can't add new words to the lattice. We find that although adapting the RNNLM on the transcript improves some test sets by a small amount, it degrades others, leaving the average WER close to the original one. We also find that adapting on the oracle transcripts is worse than adapting on ground truth, so adapting on transcripts that have errors doesn't help teaching the RNNLM to deal with errors in the lattice.  

The gains we observe with EMBR training are consistent across all test sets, and it appears that the more effective the RNNLM is at reducing the WER, the more EMBR has an impact. We reported the relative WER reduction brought by the RNNLM and RNNLM+EMBR, and we observe that on average the number of errors fixed by the RNNLM increases by almost 25\% after EMBR training, going from 7.3\% to 9.1\%.

In Table \ref{table:errors} we display the breakdown of the errors of the models on the test sets. We can see that, although EMBR increases the number of insertions and substitutions, it lowers the number of deletions by a greater margin. This goes in line with the findings of \cite{Huang}, and supports our intuition that PPL-trained LMs tend to prefer shorter sentences.
\begin{table}[!htpb]
\centering
\begin{tabular}{c|ccc}
\toprule
\rule[-1ex]{0pt}{2.5ex} Model & Insertions & Deletions & Substitutions \\ 
\midrule
\rule[-1ex]{0pt}{2.5ex} oracle & 1794 & 22370 & 10303 \\ 
\midrule
\rule[-1ex]{0pt}{2.5ex} 4-gram only & 5213 & 29968 & 23311 \\
\rule[-1ex]{0pt}{2.5ex} RNNLM & \textbf{4866} & 28911 & \textbf{20417} \\ 
\rule[-1ex]{0pt}{2.5ex} RNNLM + EMBR & 5147 & \textbf{27609} & 20504 \\  
\bottomrule
\end{tabular} 
\caption{Errors made by the models across all the test sets.}
\label{table:errors}
\end{table}

One drawback of our method is that our implementation is slow, with each epoch taking several days on a Titan X GPU. As shown in Figure \ref{fig:train_test}, it's unclear if the model has converged after 3 epochs so improvements to the implementation could yield further gains.

\section{Conclusion and future work}
\label{sec:conclusion}
We presented a method to fine-tune an RNNLM discriminatively on lattice data by minimising the expected word error rate, and showed that it was effective at reducing the WER on diverse test sets, unlike adapting on transcripts, meaning that the RNNLM doesn't only learn the language used in the training data, but also learns to give scores that are more likely to reduce the WER. The proposed EMBR training increases the relative gain from rescoring with the RNNLM rather than the 4-gram alone from 7.3\% to 9.1\%. This corresponds to a 1.9\% relative improvement in WER compared to a non fine-tuned RNNLM.  

In future we will study the impact of using different WER approximations in the lattice as noted in Section \ref{ssec:computation}. We will also optimise our implementation of the rescoring and forward-backward algorithm to make it more efficient to train, enabling us to further investigate the effect of large scale training with the proposed criterion.

\vfill\pagebreak

\bibliographystyle{IEEEbib}
\bibliography{refs,misc}

\begin{thebibliography}{10}

\bibitem{Yu:2014:ASR:2695502}
Dong Yu and Li~Deng,
\newblock {\em Automatic Speech Recognition: A Deep Learning Approach},
\newblock Springer Publishing Company, Incorporated, 2014.

\bibitem{Mohri2008}
Mehryar Mohri, Fernando Pereira, and Michael Riley,
\newblock ``{Speech Recognition with Weighted Finite-State Transducers},''
\newblock in {\em Springer Handbook of Speech Processing}, pp. 559--584.
  Springer Berlin Heidelberg, Berlin, Heidelberg, 2008.

\bibitem{Klakow2002}
Dietrich Klakow and Jochen Peters,
\newblock ``{Testing the correlation of word error rate and perplexity},''
\newblock {\em Speech Communication}, vol. 38, no. 1-2, pp. 19--28, sep 2002.

\bibitem{Huang}
Jiaji Huang, Yi~Li, Wei Ping, and Liang Huang,
\newblock ``{Large Margin Neural Language Model},''
\newblock Tech. {R}ep.

\bibitem{Povey2013}
Daniel Povey and Karel Vesel{\'{y}},
\newblock ``{Sequence-discriminative training of deep neural networks},''
\newblock {\em Interspeech}, , no. 1, pp. 3--7, 2013.

\bibitem{Povey2016}
Daniel Povey, Vijayaditya Peddinti, Daniel Galvez, Pegah Ghahremani, Vimal
  Manohar, Xingyu Na, Yiming Wang, and Sanjeev Khudanpur,
\newblock ``{Purely Sequence-Trained Neural Networks for ASR Based on
  Lattice-Free MMI},''
\newblock sep 2016, pp. 2751--2755.

\bibitem{Shannon2017}
Matt Shannon,
\newblock ``{Optimizing expected word error rate via sampling for speech
  recognition},''
\newblock jun 2017.

\bibitem{Dikici2013}
Erin{\c{c}} Dikici, Student Member, Murat Semerci, Murat Sara{\c{c}}lar, Ethem
  Alpaydın, and Senior Member,
\newblock ``{Classification and Ranking Approaches to Discriminative Language
  Modeling for ASR},''
\newblock {\em IEEE TRANSACTIONS ON AUDIO, SPEECH, AND LANGUAGE PROCESSING},
  vol. 21, no. 2, pp. 291, 2013.

\bibitem{Tachioka2015}
Yuuki Tachioka and Shinji Watanabe,
\newblock ``{Discriminative method for recurrent neural network language
  models},''
\newblock in {\em ICASSP, IEEE International Conference on Acoustics, Speech
  and Signal Processing - Proceedings}. apr 2015, vol. 2015-Augus, pp.
  5386--5390, IEEE.

\bibitem{Arora2016a}
Kushal Arora and Anand Rangarajan,
\newblock ``{Contrastive Entropy: A new evaluation metric for unnormalized
  language models},''
\newblock jan 2016.

\bibitem{Heigold2005}
G.~Heigold, W.~Macherey, R.~Schluter, and H.~Ney,
\newblock ``{Minimum exact word error training},''
\newblock in {\em IEEE Workshop on Automatic Speech Recognition and
  Understanding, 2005.} 2005, pp. 186--190, IEEE.

\bibitem{VanDalen2015}
Rogier~C {Van Dalen} and Mark J~F Gales,
\newblock ``{Annotating large lattices with the exact word error},''
\newblock Tech. {R}ep., 2015.

\bibitem{Prabhavalkar2017}
Rohit Prabhavalkar, Tara~N. Sainath, Yonghui Wu, Patrick Nguyen, Zhifeng Chen,
  Chung-Cheng Chiu, and Anjuli Kannan,
\newblock ``{Minimum Word Error Rate Training for Attention-based
  Sequence-to-Sequence Models},''
\newblock dec 2017.

\bibitem{Povey2002}
D.~Povey and P.C. Woodland,
\newblock ``{Minimum Phone Error and I-smoothing for improved discriminative
  training},''
\newblock in {\em IEEE International Conference on Acoustics Speech and Signal
  Processing}. may 2002, pp. I--105--I--108, IEEE.

\bibitem{Liu2016}
Xunying Liu, Xie Chen, Yongqiang Wang, Mark J.~F. Gales, and Philip~C.
  Woodland,
\newblock ``{Two Efficient Lattice Rescoring Methods Using Recurrent Neural
  Network Language Models},''
\newblock {\em IEEE/ACM Transactions on Audio, Speech, and Language
  Processing}, vol. 24, no. 8, pp. 1438--1449, aug 2016.

\bibitem{Kumar2017}
Shankar Kumar, Michael Nirschl, Daniel Holtmann-Rice, Hank Liao,
  Ananda~Theertha Suresh, and Felix Yu,
\newblock ``{Lattice Rescoring Strategies for Long Short Term Memory Language
  Models in Speech Recognition},''
\newblock nov 2017.

\bibitem{Xu2018}
Hainan Xu, Tongfei Chen, Dongji Gao, Yiming Wang, Ke~Li, Nagendra Goel, Yishay
  Carmiel, Daniel Povey, and Sanjeev Khudanpur,
\newblock ``{A Pruned Rnnlm Lattice-Rescoring Algorithm for Automatic Speech
  Recognition},''
\newblock in {\em 2018 IEEE International Conference on Acoustics, Speech and
  Signal Processing (ICASSP)}. apr 2018, pp. 5929--5933, IEEE.

\bibitem{Mnih2013}
Andriy Mnih and Koray Kavukcuoglu,
\newblock ``{Learning word embeddings efficiently with noise-contrastive
  estimation},'' 2013.

\bibitem{Williams2015}
Will Williams, Niranjani Prasad, David Mrva, Tom Ash, and Tony Robinson,
\newblock ``{Scaling Recurrent Neural Network Language Models},''
\newblock feb 2015.

\bibitem{Ko2017}
Tom Ko, Vijayaditya Peddinti, Daniel Povey, Michael~L. Seltzer, and Sanjeev
  Khudanpur,
\newblock ``{A study on data augmentation of reverberant speech for robust
  speech recognition},''
\newblock in {\em 2017 IEEE International Conference on Acoustics, Speech and
  Signal Processing (ICASSP)}. mar 2017, pp. 5220--5224, IEEE.

\bibitem{pykaldi}
Doğan Can, Victor~R. Martinez, Pavlos Papadopoulos, and Shrikanth~S.
  Narayanan,
\newblock ``Pykaldi: A python wrapper for kaldi,''
\newblock in {\em Acoustics, Speech and Signal Processing (ICASSP), 2018 IEEE
  International Conference on}. IEEE, 2018.

\end{thebibliography}

\end{document}